\documentclass[conference,comsoc]{IEEEtran}
\usepackage{cite}
\ifCLASSINFOpdf
\usepackage[pdftex]{graphicx}
\else
\usepackage[dvips]{graphicx}
\fi
\usepackage[cmex10]{amsmath}
\usepackage{algorithmic}
\usepackage{array}
\ifCLASSOPTIONcompsoc
\usepackage[caption=false,font=normalsize,labelfont=sf,textfont=sf]{subfig}
\else
\usepackage[caption=false,font=footnotesize]{subfig}
\fi
\usepackage{dblfloatfix}
\usepackage{booktabs}
\usepackage{url}
\usepackage{float}
\usepackage{multirow}
\usepackage{enumerate}
\usepackage{amsthm}
\usepackage{nidanfloat}
\usepackage{threeparttable}
\newtheoremstyle{mystyle}
{}
{}
{\itshape}
{}
{\bfseries}
{.}
{ }
{}
\theoremstyle{mystyle}

\newtheorem{remark}{Remark}

\newenvironment{talign*}
{\csname align*\endcsname}
{\endalign}
\usepackage{arydshln}
\usepackage[colorinlistoftodos]{todonotes}
\usepackage[normalem]{ulem}
\usepackage{hyperref}
\begin{document}
	%
	\title{Zero-bias Deep Neural Network for Quickest RF Signal Surveillance\\
		\thanks{978-1-6654-4331-9/21/\$31.00 ©2021 IEEE}}
	
	
	
	
	\author{%
		Yongxin Liu$^{1}$, Yingjie Chen$^{2}$, Jian Wang$^{3}$, Shuteng Niu$^{4}$, Dahai Liu$^{3}$, Houbing Song$^{3}$\\
		$^{1}$Auburn University at Montgomery, Montgomery, AL 36117 USA\\
		$^{2}$Qingdao University, Qingdao, Shandong 266071 China\\
		$^{3}$Embry-Riddle Aeronautical University, Daytona Beach, FL 32114 USA\\
		$^{4}$Bowling Green State University, Bowling Green, OH 43403 USA\\
		
		$^{1}$yongxin.liu@aum.edu,
		$^{2}$2018207163@qdu.edu.cn, $^{3}$wangj1@my.erau.edu, dahai.liu@erau.edu, h.song@ieee.org\\
		$^{4}$sniu@bgsu.edu
	}

	\markboth{IEEE Internet of Things Journal,~Vol.~11, No.~4, May~2021}%
	{Shell \MakeLowercase{\textit{et al.}}: Bare Demo of IEEEtran.cls for Journals}
	\IEEEtitleabstractindextext{%
		\begin{abstract}
			The Internet of Things (IoT) is reshaping modern society by allowing a decent number of RF devices to connect and share information through RF channels. However, such an open nature also brings obstacles to surveillance. For alleviation, a surveillance oracle, or a cognitive communication entity needs to identify and confirm the appearance of known or unknown signal sources in real-time. In this paper, we provide a deep learning framework for RF signal surveillance. Specifically, we jointly integrate the Deep Neural Networks (DNNs) and Quickest Detection (QD) to form a sequential signal surveillance scheme. We first analyze the latent space characteristic of neural network classification models, and then we leverage the response characteristics of DNN classifiers and propose a novel method to transform existing DNN classifiers into performance-assured binary abnormality detectors. In this way, we seamlessly integrate the DNNs with the parametric quickest detection. Finally, we propose an enhanced Elastic Weight Consolidation (EWC) algorithm with better numerical stability for DNNs in signal surveillance systems to evolve incrementally, we demonstrate that the zero-bias DNN is superior to regular DNN models considering incremental learning and decision fairness. We evaluated the proposed framework using real signal datasets and we believe this framework is helpful in developing a trustworthy IoT ecosystem.
		\end{abstract}
		
	}

	\IEEEoverridecommandlockouts

	\maketitle
	
	\IEEEdisplaynontitleabstractindextext

	%
	\IEEEpeerreviewmaketitle

	\section{Introduction}
	%
	%
	%
	%
	
	The Internet of Things (IoT) is providing applications and services that would otherwise not be possible \cite{7406686,song2016cyber}. Intelligent decision making is of great significance in IoT \cite{9345787}. A typical way to implement smart decision functionality in IoT is by integrating learning-enabled components through Deep Learning (DL) and Deep Neural Networks (DNNs). One typical application of DNNs in IoT is the RF signal surveillance to either identify device type or modulation schemes \cite{8897627,9500733,9425491}.
	
	Although DL and DNNs have been applied in recognition of RF signals for device identification \cite{liu2021machine} and event surveillance \cite{wang2021counter}, applying DNNs in safety-critical systems requiring assured performance is still controversial. Firstly, DNNs perform well on specifying known subjects but cannot distinguish abnormalities. Abnormal signals, such as those from unauthorized signal sources, are required to be identified accurately rather than being erroneously classified into the most likely known ones \cite{jiang2019uncertainty}. Secondly, DNN related systems lack the characteristics of timeliness assurability, while applications in safety-critical systems require making accurate decisions with both theoretical assured minimum latency and pre-defined false alarm constraints. The two obstacles impede the deployment of DL and DNNs in IoT of safety-critical systems. 
	
	For unknown event detection, the most intuitive way is to use statistical models to generate likelihood metrics and then use thresholds to distinguish whether an input is within the learned knowledge domain. However, selecting features from data and specifying statistical metrics can also be time-consuming. Existing works use deep autoencoders or Generative Adversarial Networks (GANs) as well as reconstruction loss to measure whether an input is from some known domain. However, training deep autoencoders or GAN models is more computationally expensive. Moreover, autoencoders or GAN models do not guarantee to respond with constrained false alarms or predictable behaviors \cite{perera2017efficient}.
	
	The problem of timeliness assurability has been widely discussed in Quickest Detection (QD) algorithms. QD algorithms are widely applied for detecting the abrupt change of statistical parameters with the lowest latency under given false alarm constraints. Existing Quickest Detection (QD) algorithms can detect changes with minimum latency under constrained false alarms. They are neither sufficient in handling high dimension inputs nor can they provide mathematically assured performance. Even though there are some methods to integrate quickest detection with DNNs, the performances of the connected systems are only measurable but not strictly assurable. We have to claim that there is a gap between machine learning and QD. 
	
	In this paper, we utilize the enhanced deep learning framework based on our previous work \cite{liu2020zero}, the zero-bias DNN model, and significantly enhance it for quickest and reliable classification of wireless signals. In this DL framework, we facilitate DNNs with both explainable behaviors in distinguishing known or abnormal inputs. Besides, with minimum latency under certain false alarm constraints. Furthermore, our solution efficiently transforms existing DNNs into abnormality detectors with predictable performance. The effectiveness of the proposed framework in handling massive signal recognition has been demonstrated. Our contributions are as follows:
	\begin{itemize}
		\item We explored the latent space characteristics of DNNs and discovered a novel method to efficiently transfer existing DNN classifiers into DNN abnormality detectors with adaptive decision boundaries. 
		\item We provide a more stable Elastic Weight Consolidation (EWC) algorithm and show that zero-bias DNNs are more reliable than regular DNNs during incremental learning.
		\item We combine our zero-bias DNN model with the parametric Quickest Change Detection theory, and our validation on massive real signal detection demonstrates the effectiveness of our integral solution.
	\end{itemize}
	
	Our research offers a solution to the accurate identification of RF signals with an assured performance, thus useful in promoting trustworthy IoT and deepening the understanding of deep neural networks. Besides, the successful integration of the neural network and QD enables the move from IoT to real-time control. 
	
	The remainder of this paper is organized as follows: A literature review of related works is presented in Section~\ref{sectRW}. We present the methodology in Section~\ref{sectMM}. Performance evaluation is presented in Section~\ref{sectEED} with conclusions in Section~\ref{sectCC}.
	
	\section{Related Work}
	\label{sectRW}
	
	Real-time event detection is a critical function in safety-critical IoT. From the perspective of input data, we may categorize them into single-shot and sequential detection paradigms. In single-shot detection \cite{liu2020zero}, event detections are performed per observation, and the past data will not be retained for future use. In contrast, the sequential detection paradigm allows accumulating information from past observations \cite{perera2017efficient}. 
	
	\subsection{Single-shot unknown event detection in DNN}
	
	Event detection plays an increasingly important role in safety-critical and latency-constrained IoT, e.g., the aviation communication system. Detecting known events are straightforward while detecting abnormal or unseen events is more difficult. 
	
	A critical problem for DL-enabled signal identification systems is that classifiers only recognize pretrained data but can not deal with abnormal or unknown data. From the perspective of DL, this issue is categorized as the Open Set Recognition \cite{scheirer2012toward,bendale2016towards} problem. An intuitive solution is to model the distribution in the latent space. In \cite{wong2018clustering}, the authors first trained a CNN model with a Softmax output on known data. They then remove the Softmax layer and turn the neural network into a nonlinear feature extractor. Finally, they use the DBSCAN algorithm to perform cluster analysis on the remapped features and show that the method has the potential of detecting a limited number of unknown classes. In \cite{shi2019deep}, the authors provide two methods to deal with abnormalities: i) Reuse trained convolutional layers to transform inputs to feature vectors, and then use Mahalanobis distance to judge the outliers. ii) Reuse the pretrained convolutional layers to transform signals to feature vectors and then perform k-means (k = 2) clustering to discover the groups of outliers. Another approach is to leverage the characteristics of generative models. In \cite{roy2019rfal}, the authors use the Generative Adversarial Network (GAN) to generate highly realistic fake data. Then they exploit the discriminator network to distinguish whether an input is from an abnormal source. 
	
	\subsection{Sequential event detection}

	From the perspective of the stochastic process, a wireless communication system in different states can be described by distributions with measurable statistical properties \cite{lai2008quickest}. Therefore, transitions within states cause the change of those properties. The quickest detection aims to detect the change as quickly as possible, subject to false alarm constraints \cite{poor2008quickest}. Considering whether prior observations are independent of an abnormal event's appearance, the optimization scheme can be defined in different forms as in \cite{johnson2017detecting}. We can also categorize the quickest event detection methods into two branches: a) detecting events with known post-change distributions. b) detecting events with unknown post-change distributions. Generally, detecting known events is faster with the Cumulative Sum Control Chart (CUSUM) algorithm can be applied directly \cite{basseville1993detection,granjon2013cusum}. A postchange distribution may not be known in some scenarios in advance, and nonparametric strategies have to be used and bring higher latency.
	
	Quickest detection provides a performance-assured solution to detect change points (related to events) in sequential data. However, the selection of statistic metrics still depends on trial and error. We focus on real-time sequential detection of events, especially on integrating the quickest detection theory with deep learning to provide an automated and performance-assured solution to latency-constrained CPS.
	\section{Problem Definition}
	Suppose that we have a sequence of signal vectors denoted as:
	\begin{align}
		\boldsymbol{SS}=\{ \boldsymbol{S_1},\dots,\boldsymbol{S_k},\dots,\boldsymbol{S_{n}}\}    
	\end{align} 
	Suppose that some known or unknown events will occur at time $k$, our signal surveillance system is required to detect the occurrence of the known or unknown event with minimal delay. 
	
	One straightforward method is to use a DNN model $D(\cdot)$ to process $\boldsymbol{SS}$ sequentially, the goal of $D(\cdot)$ is to provide a score for each signal element to quantify whether it is from the previous known knowledge domain. From the perspective of domain adaption, feature extractors are specifically trained to fit the characteristics only within their learned tasks \cite{wang2018deep}, the task-specific knowledge domain. However, the DNN model, $D(\cdot)$, can be trivial to use. Firstly, we do not have a good method to explain or adjust the decision threshold for $D(\cdot)$. Secondly, $D(\cdot)$ can generate false alarms or encounter miss detection. We need to find a sequential detection scheme that can aggregate evidence sequentially and provide minimal detection latency. Finally, if the signal surveillance system is required to evolve incrementally, $D(\cdot)$ needs to be retrained frequently with large overhead as new data are emerging incrementally. Therefore, we need to develop a new DL paradigm that: a) enables explainable and reliable event detection. b) being able to learn incrementally and adapt to operational variations.
	
	\section{Methodology}
	\label{sectMM}
	\subsection{The zero-bias neural network}
	
	\begin{figure}[]
		\centering
		\includegraphics[width=0.9\linewidth]{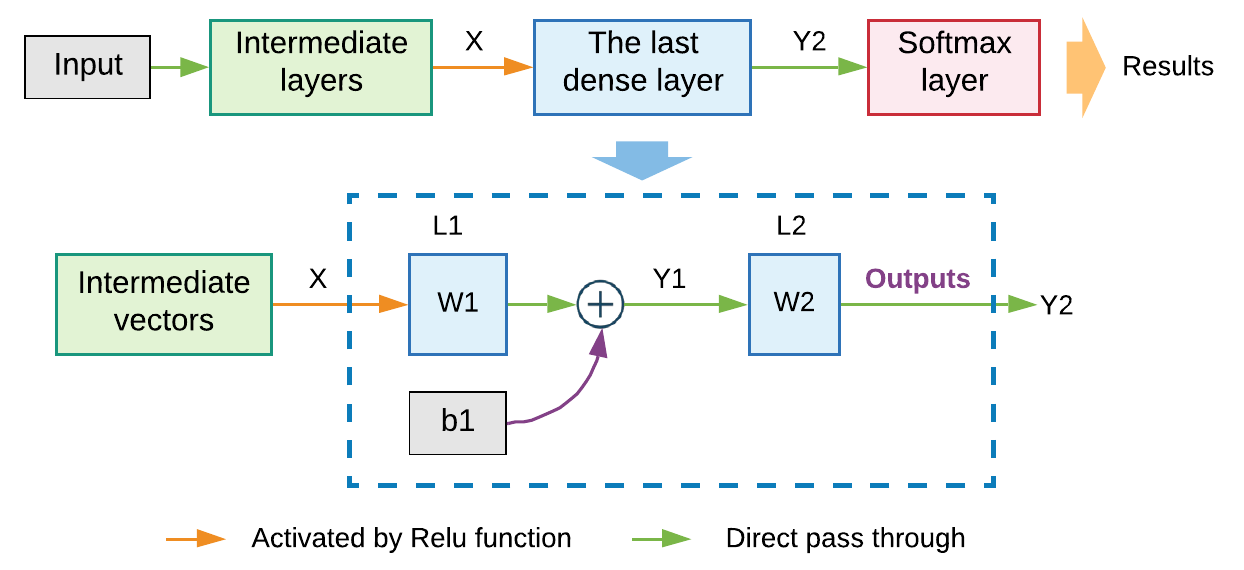}
		\caption{Data flow of zero-bias deep neural networks.}
		\label{figZerobiasDense}
	\end{figure}
	We have discovered that the last dense layer of a DNN classifier performs the nearest neighbor matching with biases and preferabilities using cosine similarity. We also show that a DNN classifier's accuracy will not be impaired if we replace its last dense layer with a zero-bias dense layer \cite{liu2020zero}, in which the decision biases and preferabilities are eliminated. We can denote its mechanism as (also in Figure~\ref{figZerobiasDense}):
	\textcolor{black}{
		\begin{align}
			\label{eqZeroBiasDenseLayer}
			\notag\boldsymbol{Y_1}(\boldsymbol{X}) &= \boldsymbol{W_0}\boldsymbol{X} + \boldsymbol{b}\\
			\boldsymbol{Y_2}(\boldsymbol{X}) &=  cosDistance(\boldsymbol{Y_1}, \boldsymbol{W_1})
		\end{align}
	}
	where $\boldsymbol{X}$ is the output of the prior convolution layers, a.k.a., feature vectors. $\boldsymbol{X}$ is an $N_0$-D vector, where $N_0$ denotes the number of features. $\boldsymbol{W}_0$ is an $N_0$ by $C$ matrix where $C$ denotes the feature dimension in the latent space, which equals the number of classes. $\boldsymbol{W_1}$ is a matrix to store fingerprints of different classes, namely the similarity matching layer, and it is a $C$ by $C$ square matrix. Please be noted that in $\boldsymbol{W_1}$, each row represents a fingerprint of the corresponding class while in $\boldsymbol{Y}_1$ each column represents a feature vector in the latent space. In short, the last dense layer is spitted into two layers, $L_1$ for feature embedding and $L_2$ for vector scaling and similarity matching. We have the first remark:
	\begin{remark}{Latent space of neural networks: }
		The latent space of a neural network for the final classification is a unit hypersphere surface. We define it as the classification hypersphere surface.
	\end{remark}
	We have proved that the classification comparison in a regular neural network is the angular matching with class-specific biases and weights, while in a zero-bias neural network, the biases are eliminated, and the weights are equalized to one. We assume that decisions can not be made according to biases in safety-critical systems. Besides that, our previous work has demonstrated that such a modification will not impair the classification performance of DNNs\cite{liu2020zero}.
	\begin{figure}[h]
		\centering
		\includegraphics[width=0.9\linewidth]{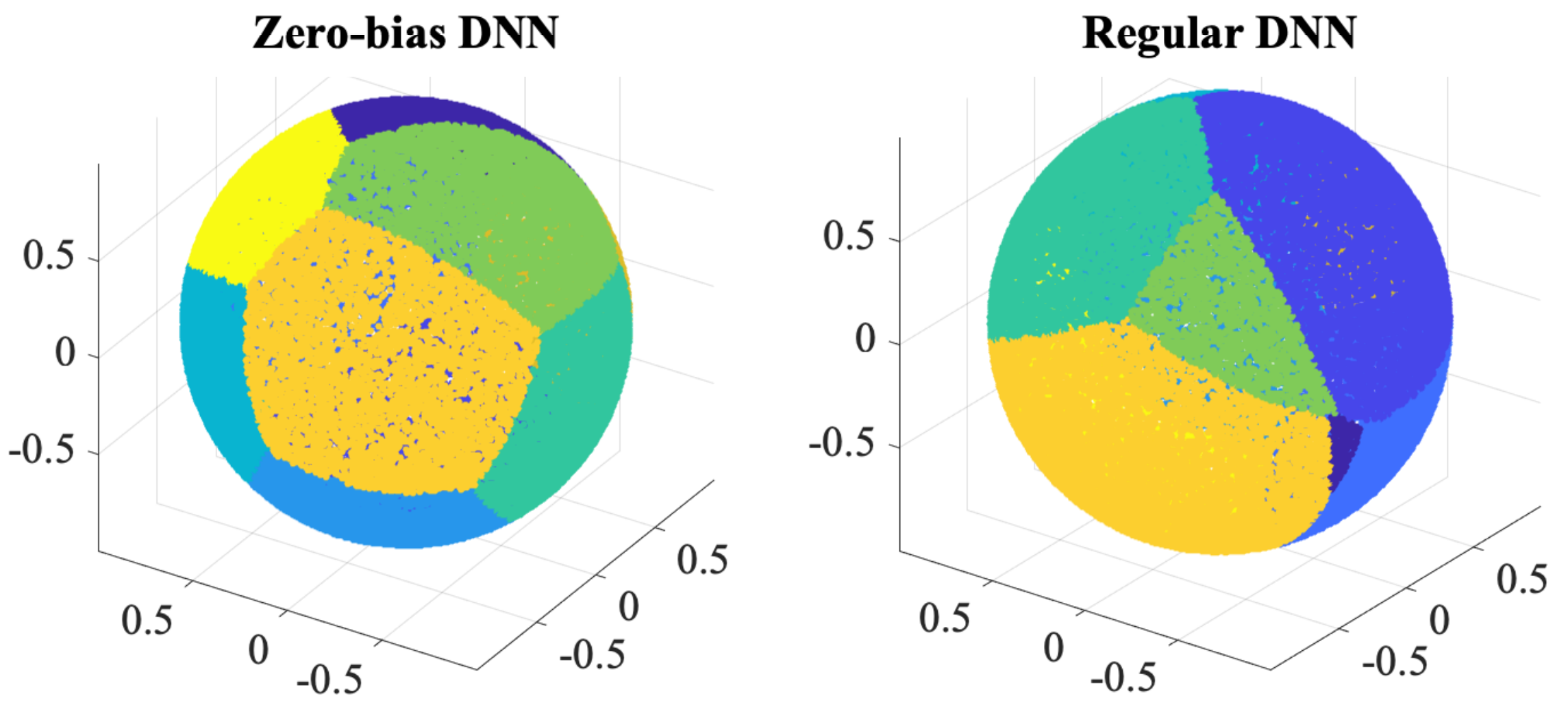}
		\caption{Associate region of classes on MNIST dataset. Data are mapped into a 3D space using t-SNE. Colors represent different classes. Code available at \url{https://github.com/pcwhy/NeuralDBVis}}
		\label{figRegionVisualization}
	\end{figure}
	
	To demonstrate this characteristic, we use a hand-written digit classification model as in \cite{matlabMinst}, and we convert it to a zero-bias neural network and retrain it. Next, we then generate random points that uniformly cover the classification hypersphere surfaces of the two models and associate the random points with their nearest class fingerprint. Finally, we use the t-SNE \cite{maaten2008visualizing} algorithm to remap the class fingerprints into a 3D hypersphere and visualize the association region of each class as in Figure~\ref{figRegionVisualization}.
	
	\begin{figure}[h]
		\centering
		\includegraphics[width=0.9\linewidth]{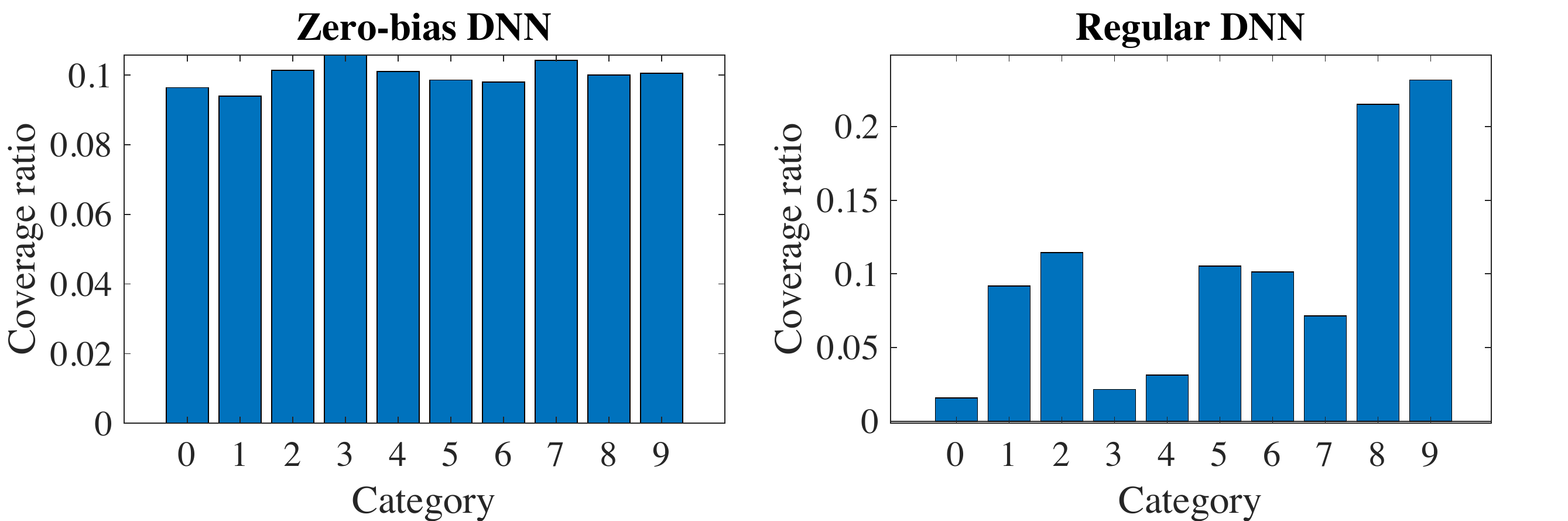}
		\caption{Compare of classification hypersphere coverage ratio of on MNIST dataset. }
		\label{figCompCoverageRatio}
	\end{figure}
	
	As depicted, in the zero-bias DNN, the classification hypersphere is uniformly divided into subregions for different classes. The subregions are not uniform when it is in the regular DNN. A more explicit numerical comparison is given in Figure~\ref{figCompCoverageRatio}. The DNN model in signal surveillance system needs to treat the input signals without biases and preferences, and thus the zero-bias neural networks can perform much better than the regular neural networks in distinguishing abnormalities (identifying the unknown input data as summarized in Table~\ref{tabAbnormalityDetectorComp}). Although one-class SVM is slightly better, it requires an optimal threshold case by case.
	
	\begin{table}[h]
		\centering
		\caption{Performance of abnormality detectors.}
		\label{tabAbnormalityDetectorComp}
		\resizebox{.4\textwidth}{!}{
			
			\begin{threeparttable}[b]
				\begin{tabular}{@{}ccccc@{}}
					\toprule
					Metric & \multicolumn{1}{l}{One-class SVM}\tnote{1} & \multicolumn{1}{l}{Zero-bias DNN}\tnote{1} & \multicolumn{1}{l}{Regular DNN}\tnote{1} \\ \midrule
					\begin{tabular}[c]{@{}c@{}}False\\ Positive\end{tabular} & 0.19 & \textbf{0.2} & 0.2 \\ \midrule
					\begin{tabular}[c]{@{}c@{}}False\\ Negative\end{tabular} & 0.05 & \textbf{0.05} & 0.28 \\ \bottomrule
				\end{tabular}
				\begin{tablenotes}
					\item [1] We set a threshold value on the maximum matching score of each input, and the threshold is set according to the maximum margin of separation as in \cite{liu2020zero}.
				\end{tablenotes}
			\end{threeparttable}
		}
	\end{table}
	Therefore, we believe the zero-bias neural network is will be a better tool to enable the intelligent surveillance of RF signals in IoT.

	\subsection{Zero-bias neural network for unsupervised and adaptive abnormality detection}
	\label{sectZbAdaptiveAbnormality}
\begin{figure}[h]
		\centering
		\includegraphics[width=0.95\linewidth]{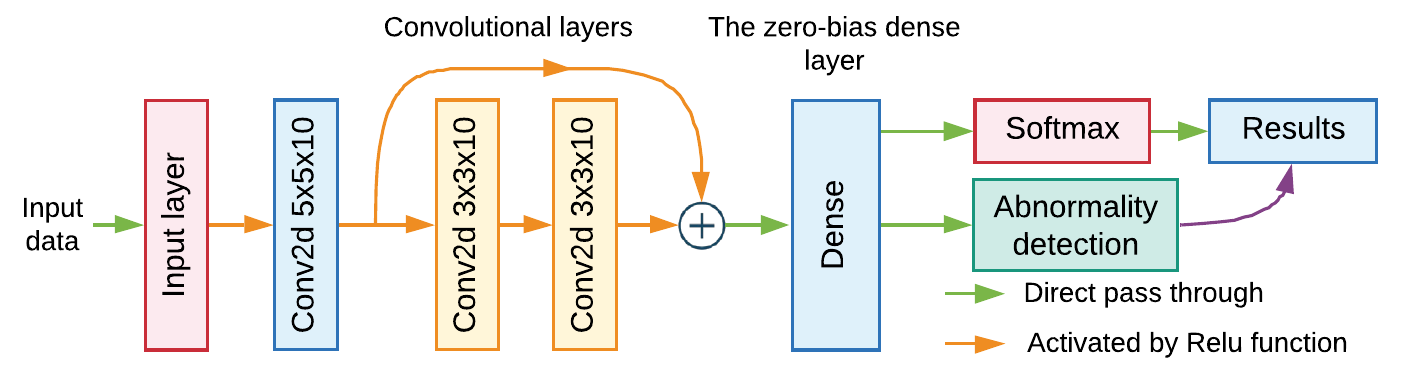}
		\caption{Deep neural network architecture \cite{liu2020deep}.}
		\label{figLearningArch}
	\end{figure}

	The zero-bias neural network compares class fingerprints and mapped data in the feature space is fair without bias and weights. Naturally, we could assume that:
	\begin{remark}
		For each fingerprint in the classification hypersphere, a cut-off cosine similarity value will separate the feature vectors of known and abnormal data. We define this value as the cut-off distance of this fingerprint.
	\end{remark}
	To verify this assumption, we use an aircraft ADS-B signal dataset \cite{gt9v-kz32-20} with the corresponding zero-bias DNN signal emitter identification model in Figure~\ref{figLearningArch}. 
	
\begin{figure}
		\centering
		\includegraphics[width=0.9\linewidth]{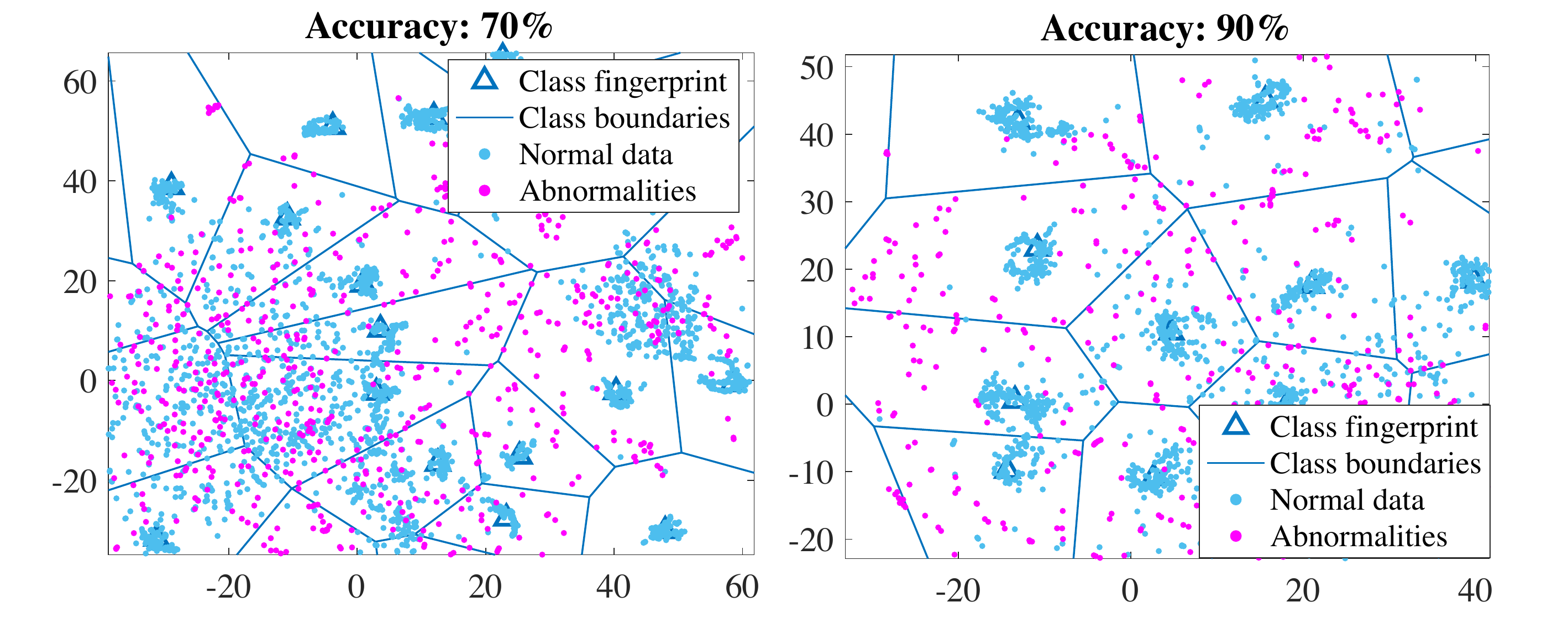}
		\caption{Class fingerprints, feature vectors of known and abnormal data in the classification hypersphere in the zero-bias neural network for signal identification}
		\label{figComp70vs90}
	\end{figure}

\begin{figure}
		\centering
		\includegraphics[width=0.6\linewidth]{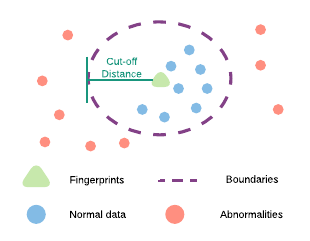}
		\caption{Class cut-off distance for distinguishing known and abnormal feature vectors in the classification hypersphere of zero-bias DNN.}
		\label{figClassBoundary}
	\end{figure}

	We first train the network in different stages. We then use the t-SNE algorithm to visualize the class fingerprints, feature vectors from known and abnormal data in the classification hypersphere as in Figure~\ref{figComp70vs90}. We can find several important features:
	\begin{itemize}
		\item The sizes of clusters for different classes are gradually becoming smaller.
		\item The abnormalities are gradually becoming more distinctively separated from the known data. We can depict the relation of feature vectors from regular data and abnormalities as in Figure~\ref{figClassBoundary}.
		\item The abnormalities (signals from unknown RF emitters) distribute randomly throughout the classification hypersphere.
	\end{itemize}
	
	For a given DNN model with the zero-bias dense layer, we follow these steps to find the cut-off distance of each class fingerprint:
	\begin{enumerate}[\textbf{Step} 1:]
		\item The training set is utilized to learn the boundaries of known classes while the validation set will be mixed with abnormal data ($\boldsymbol{A_0}$) to measure the performance of the converted abnormality detector.
		\item We pass accurately classified data of $i$th known class from the training set, denoted as $\boldsymbol{KX}_i$, through layers of the DNN model and obtain the compressed feature vectors before fingerprint matching, denoted as:
		\begin{align}
			\boldsymbol{Y}_1[F_{n-1}(\boldsymbol{KX}_i)] = \boldsymbol{W}_0 F_{n-1}(\boldsymbol{KX}_i) + \boldsymbol{b}
		\end{align}
		Where $\boldsymbol{W}_0$ and $\boldsymbol{b}$ are defined in Equation~\ref{eqZeroBiasDenseLayer}, $F(\cdot)_{n-1}$ denotes all network layers before the fingerprint matching. $\boldsymbol{Y}_1[F_{n-1}(\boldsymbol{KX}_i)]$ denotes feature vectors of accurately classified data in $\boldsymbol{KX}_i$.
		\item Calculate the centroid $\boldsymbol{c}_0^i$ of $\boldsymbol{KX}_i$ as:
		\begin{align}
			\boldsymbol{c}_0^i &= mean( \boldsymbol{Y}_1[F_{n-1}(\boldsymbol{KX}_i)])
		\end{align}
		\item Calculate all the cosine distances between the $\boldsymbol{c}_0^i$ to all accurately classified feature vectors. We then use the greatest cosine distance value as the cut-off distance, $CO_i$, for the $i$th known class.
		
		\item Abnormality detection using cut-off boundaries on input data $\boldsymbol{X})$ is formally defined as:
		\begin{align}
			\hspace*{\dimexpr-\leftmargini}
			D(\boldsymbol{X}) = 
			\begin{cases}
				1~~\exists~i,~ cosineDistance[\boldsymbol{Y}_1,\boldsymbol{c}_0^i] \leq CO_i\\
				0~~\text{Otherwise}
			\end{cases}
		\end{align}
	\end{enumerate}
	These steps convert a zero-bias DNN into an abnormality detector with binary output. In this binary abnormality detector, we do not need to specifically adjust its decision threshold compared with our previous method in \cite{liu2020zero}. According to our observation, as long as the zero-bias DNN is well-trained before conversion, very few anomaly data may have a cosine distance less than the cut-off distance.
	
	From the perspective of signal surveillance, we may randomly encounter known or abnormal signals. Therefore, the output of the binary abnormality detector can also be regarded as a random sequence, in which when the signals are from some known events, the output of the abnormality detector follows a Bernoulli distribution: \cite{weisstein2002bernoulli}:
	\begin{align}
		P_0(I_k) &= FPR^{I_k}(1-FPR)^{1-I_k}
	\end{align}
	where $I_k \in \{0,1\}$ is the output of the binary abnormality detector with $I_k = D(\boldsymbol{X_k})$. $FPR$ is the false positive rate of the binary abnormality detector.
	
	When we encounter some abnormal events, the output of the binary abnormality will become:
	\begin{align}
		P_1(I_k) &= (1-FNR)^{I_k}FNR^{1-I_k}\\
		&=(TPR)^{I_k}(1-TPR)^{1-I_k}    
	\end{align}
	where $TNR$ and $TPR$ are the true negative and true positive rates of the binary abnormality detector. 
	
\begin{figure}[h]
		\centering
		\includegraphics[width=0.85\linewidth]{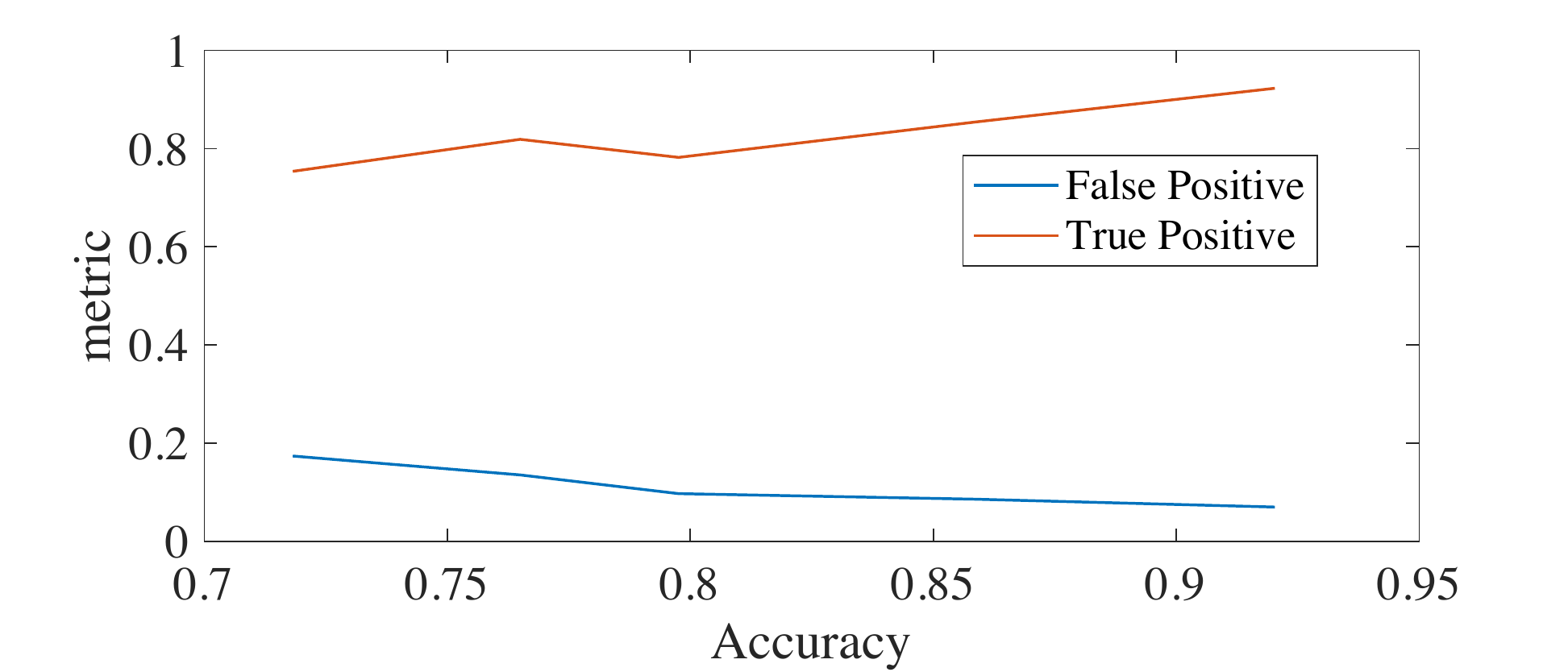}
		\caption{Performance of the converted abnormality detector.}
		\label{figPerformanceMetricAbn}
	\end{figure}

	Relations of $FPR$, $TPR$ of the binary abnormality detector, and the training accuracy of zero-bias DNN before conversion are depicted in Figure~\ref{figPerformanceMetricAbn}. These relations can be quantified using two linear models on both MNIST \cite{matlabMinst} and our ADS-B signal dataset \cite{gt9v-kz32-20,liu2020deep}. They are:
	\begin{align}
		FPR = 1 - ACC~~~R^2 = 0.85\\
		TPR = 0.2 + 0.77 \cdot ACC~~~R^2 = 0.89
	\end{align}
	Therefore, we can directly use the training accuracy of the zero-bias DNN model as an predictor for $FNR$ and $TPR$ of the converted binary abnormality detector.
	
	\subsection{Sequential event detection with zero-bias neural network}
	As we have converted the regular zero-bias DNN model into a binary abnormality detector and have formulated the behavior of this model using two Bernoulli Distributions with predictable parameters. We can then define the event detection problem as a sequential statistical test scheme using the CUSUM algorithm.
	
	First, a likelihood ratio test is employed to sequentially process the observed data at each timestamp $k$, denoted as:
	\begin{align}
		g(k) = ln(\dfrac{P_1(\boldsymbol{S}_k)}{P_0(\boldsymbol{S}_k)})
	\end{align}
	where $g(k)$ is a sufficiency metric, $P_0(\cdot)$, $P_1(\cdot)$ denotes the probabilistic density functions of abnormal and abnormal states, respectively. A constrained cumulative sum of sufficiency metrics is used as an indicator, denoted as:
	\begin{align}
		s(k) = max(0,s(k-1) + g(k))
	\end{align}
	An alarm will be sent once $s(k)$ is greater than a predefined threshold, $h_{CUSUM}$, and this alarm will indicate that some unknown events are happening. The CUSUM algorithm has been proved to provide the lowest worst-case detection latency at specific false alarm intervals \cite{xie2021sequential,granjon2013cusum}. Therefore, if our solution is applied, the detection threshold, $h_{CUSUM}$, is the only parameter that needs to be specified. 
	
	\subsection{Light-weight incremental learning algorithm for zero-bias neural network}
	\label{sectContinualLearning}
	A benefit of zero-bias DNN enabled binary abnormality detector is that incremental learning can be implemented to facilitate the model to evolve in its lifecycle. Incremental learning enables a neural network to classify new targets without needing to retrain from scratch. In our research, specific events can be learned as new classes to be recognized directly. For both zero-bias DNN and conventional DNN classifiers: 
	\begin{remark}[Incremental learning on the classification hypersphere]
		\label{rmContinualLearning}
		To enable a neural network to recognize a new class, we only need to place its class fingerprint on the classification hypersphere and fine-tune the old fingerprints' directions when necessary. 
	\end{remark}{}
	We also have: 
	\begin{itemize}
		\item For a specific new class, as long as the previous layers have extracted sufficient distinctive features, we do not need to retrain the previous layers.
		\item For new classes, we need to insert new fingerprints and then adjust the old fingerprints when necessary.
	\end{itemize}
	
	To adjust an old fingerprint, we need to identify which parameter (or dimension) is critical to the classification accuracy. According to the Elastic Weight Consolidation (EWC) \cite{kirkpatrick2017overcoming}, the Fisher Information Matrix is used to model the importance of parameters as: 
	\begin{align}
		&\boldsymbol{F_{\boldsymbol{\Omega}}}=[\dfrac{\partial log(P(\boldsymbol{X}_{CV}|\boldsymbol{\Omega}))}{\partial \boldsymbol{\Omega}}][\dfrac{\partial log(P(\boldsymbol{X}_{CV}|\boldsymbol{\Omega}))}{\partial \boldsymbol{\Omega}}]^T\notag\\
		&P(\boldsymbol{X}_{CV}|\boldsymbol{\Omega})\approx\overline{\boldsymbol{Y}_{Softmax}}(\boldsymbol{X}_{CV}|\boldsymbol{\Omega})
	\end{align}
	Where $\overline{\boldsymbol{Y}_{Softmax}}(\boldsymbol{X}_{CV}|\boldsymbol{\Omega})$ denotes the averaged outputs of Softmax layer on validation set $\boldsymbol{X}_{CV}$ given parameter set $\boldsymbol{\Omega}$, it approximates the posterior probability $P(\boldsymbol{X}_{CV}|\boldsymbol{\Omega})$. $\boldsymbol{F_{\boldsymbol{\Omega}}}$ denotes the Fisher information matrix of the current task. In our experiment, we further apply an exponential function to the Fisher Information to increase the numerical stability as:
	\begin{align}
		\label{eqFisherExp}
		\boldsymbol{F_{\boldsymbol{\Omega}}}:=\exp{(\boldsymbol{F_{\boldsymbol{\Omega}}})}
	\end{align}
	Intuitively, the importance of a parameter is equivalent to the square of its gradient with respect to the logarithm of the Softmax function. 
	
	Knowing the importance of existing parameters, we can define an integral loss function for incremental learning as:
	\begin{align}
		\label{eqContinualLearnLoss}
		F_1(\boldsymbol{\Omega})&=\dfrac{\lambda_1}{2} \sum _{i}[\boldsymbol{F_{\boldsymbol{\Omega^*}}}\cdot(\boldsymbol{\Omega}-\boldsymbol{\Omega^*})^2]\notag\\
		L(\boldsymbol{\Omega})&=( L_2(\boldsymbol{\Omega})+F_1(\boldsymbol{\Omega}))\cdot \boldsymbol{G_m}
	\end{align}
	Where $F_1(\boldsymbol{\Omega})$ denotes the Fisher Loss with respect to old tasks (a.k.a., task-1). $\boldsymbol{\Omega^*}$ denotes the loss function and model parameters on task-1. $L_2(\boldsymbol{\Omega})$ and $\boldsymbol{\Omega}$ denote the raw loss function on Task-2 and the new model parameters. $\lambda_1$ denotes the importance of task-1. Intuitively, this integral loss function additionally penalizes the change of critical parameters. $\boldsymbol{G_m}$ is a mask matrix to control which parameter is locked or unlocked. The value of each element can only be zero or one.
	
	Given a neural network trained on Task-1 ($DNN_1$), incremental learning on Task-2 is performed as follows:
	\begin{enumerate}[\textbf{Step} 1:]
		\item Store all learnable parameters of $DNN_1$ as $\boldsymbol{\Omega^*}$ and calculate their importance matrix $\boldsymbol{F_{\boldsymbol{\Omega^*}}}$.
		\item Generate the initial fingerprint of each new class by averaging their feature vectors.
		\item Concatenate initial fingerprints into the last dense layer or zero-bias dense layer.
		\item Lock the weights of previous layers and calculate the importance of parameters of old fingerprints. The importance of newly concatenated fingerprints is set to zeros; thus, we could allow them to learn freely.
		\item Use loss function as in Equation (\ref{eqContinualLearnLoss}) and a training set of Task-2 to perform network training.
	\end{enumerate}
	Notably, we do not need to retain old training data to learn a new task, and such a benefit is critical for DNN models in practical scenarios. 
	
	
	\section{Evaluation and Discussion}
	\label{sectEED}
	\subsection{Evaluation dataset}
	Our dataset is available in \cite{gt9v-kz32-20}, we use the wide-spreading signals from Automatic Dependent Surveillance-Broadcast (ADS-B) signals \cite{riddle1090}, which provides a great variety of signals from commercial aircraft's RF transponders with labels. Specifically, each aircraft use transponders at 1090MHz to broadcast its flight information to the Air Traffic Control (ATC) center. The integrity and trustworthiness of ADS-B messages are critical to aviation safety. However, the ADS-B system does not contain cryptographic identity verification mechanisms and thus is vulnerable to identity spoofing attacks. Our previous works \cite{liu2020zero,liu2020deep} have shown that the responses of the zero-bias DNN to known (learned) aircraft and unknown sources (also from unknown aircraft) can be modeled by different probability distributions. Here we define the appearance of unknown aircraft's signals as abnormal events, and we can use the framework in this paper to design a sequential event detector to aggregate warnings and identify the adversaries who use fake IDs.
	
	From the perspective of DL, the input is the raw signal collected by a Software Defined Radio Receiver (USRP B210), and the DNN is trained to identify the known aircraft through their signals. As in our previous work \cite{liu2020zero, liu2020deep}, we take the first 1024 samples from each signal record and extract pseudo-noise, magnitude-frequency, and phase-frequency features. The extracted features of each signal record are then packed into a 32 by 32 by 3 tensor. The architecture of our DNN model is depicted in Figure~\ref{figLearningArch} with a description of the dataset in Table~\ref{tabDataUsage}. After training to recognize known aircraft, the zero-bias DNN model is then converted to a binary abnormality detector as in Section~\ref{sectZbAdaptiveAbnormality}. 
	
	\begin{table}[h]
		\caption{Description of dataset}
		\label{tabDataUsage}
		\centering
		\resizebox{.4\textwidth}{!}{
			\begin{tabular}{cl}
				\toprule
				Usage & \multicolumn{1}{c}{Description} \\ \midrule
				Training & \begin{tabular}[c]{@{}l@{}}60\% of signal records from 28 aircraft.\end{tabular} \\ \midrule
				Test & \begin{tabular}[c]{@{}l@{}}40\% of signal records from 28 aircraft.\end{tabular} \\ \midrule
				Normal data & \begin{tabular}[c]{@{}l@{}}The test set.\end{tabular} \\ \midrule
				Abnormal data & \begin{tabular}[c]{@{}l@{}}Signal records from the remaining 100 aircraft.\end{tabular} \\ \bottomrule
			\end{tabular}
		}
	\end{table}
	
	\subsection{Quickest abnormal event detection}
\begin{figure}[h]
		\centering
		\includegraphics[width=0.8\linewidth]{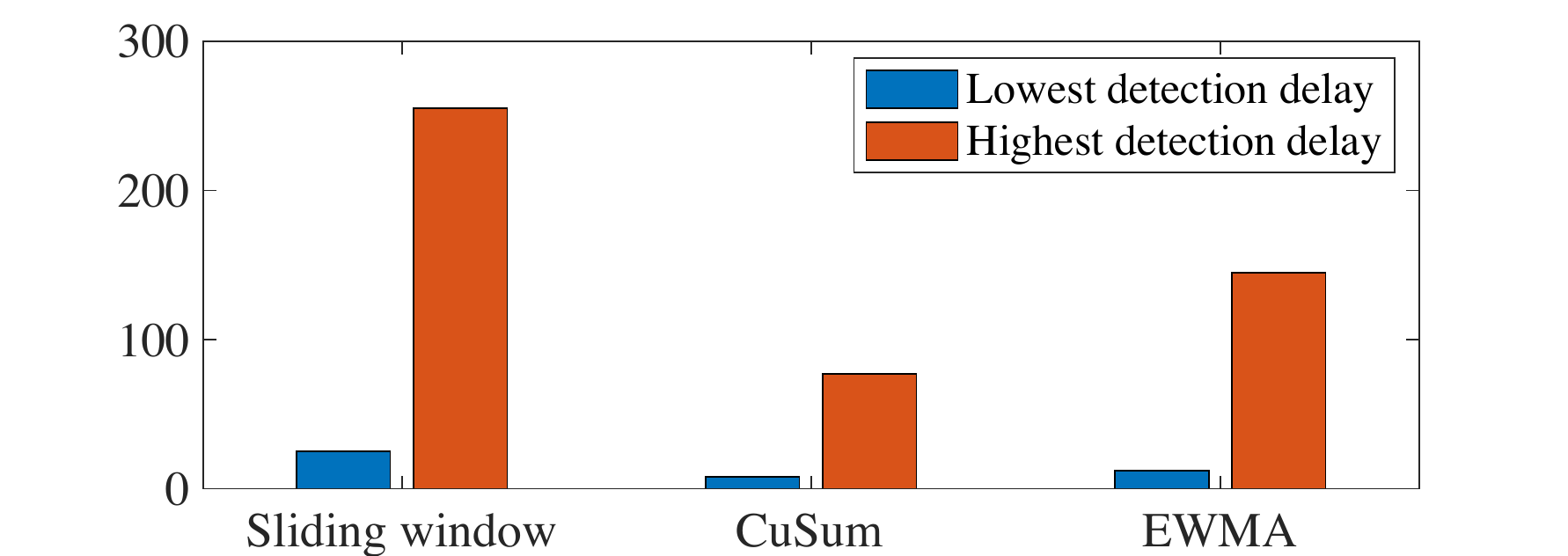}
		\caption{The best and the worst case detection delay}
		\label{figCompDetectionDelayAll}
	\end{figure}

\begin{figure}[h]
		\centering
		\includegraphics[width=0.8\linewidth]{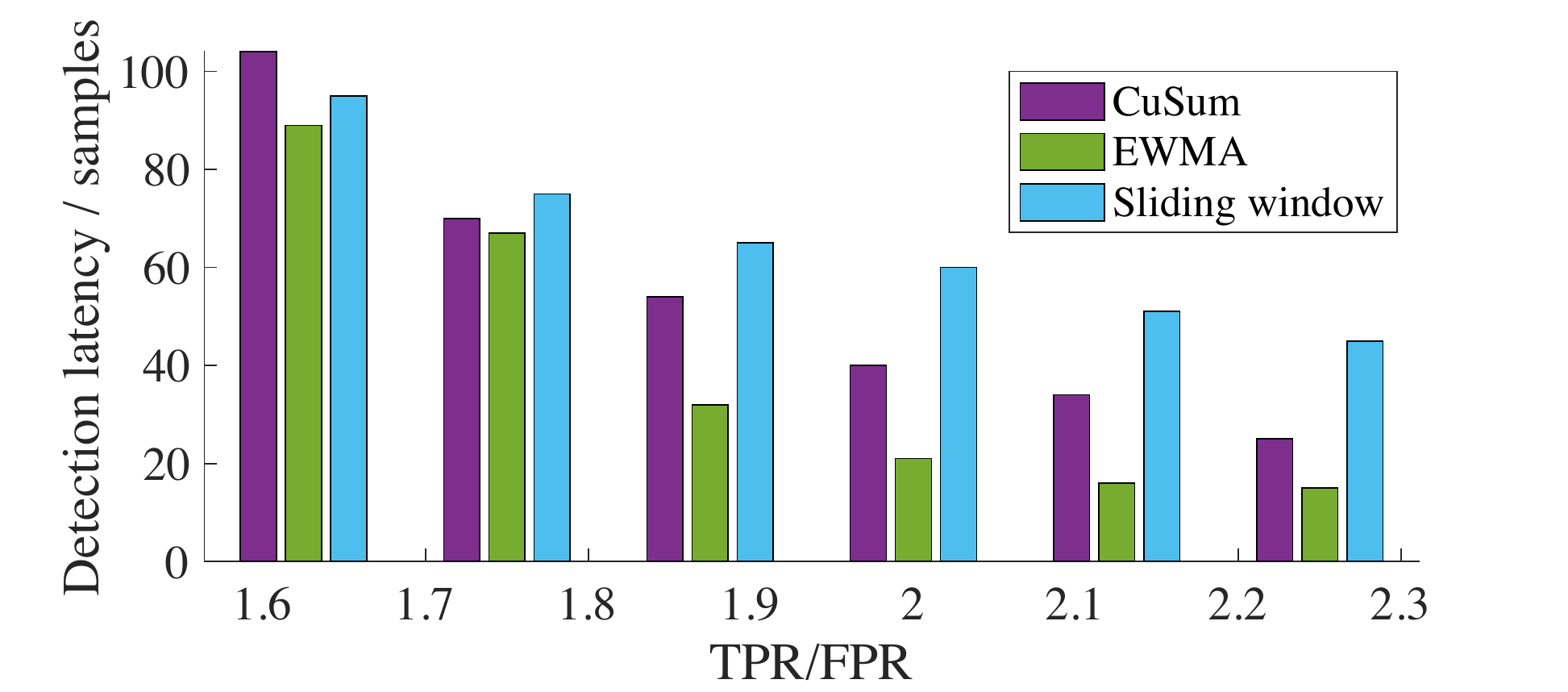}
		\caption{The averaged detection delay.}
		\label{figAvgDetectionDelayAllCases}
	\end{figure}

\begin{figure}[h]
		\centering
		\includegraphics[width=0.8\linewidth]{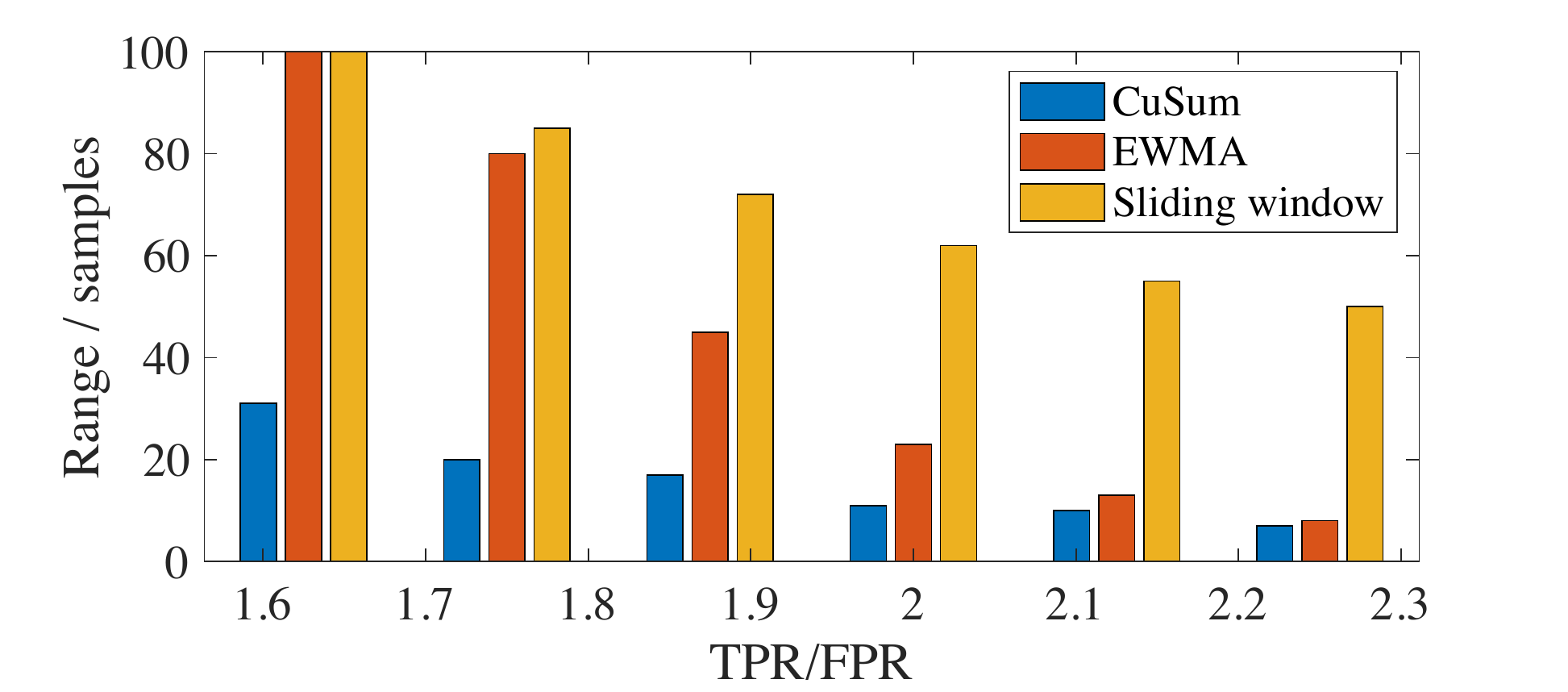}
		\caption{The ranges of detection delay.}
		\label{figCompDetectionDelayRangeAll}
	\end{figure}

	The converted binary abnormality detector can be utilized for abnormal event detection with very low latency as a result of both high true positive and low false positive rates. To further evaluate our proposed method,
	we first define a quality metric, $Q = \dfrac{TPR}{FPR}$, for the binary abnormality detector. Then, we can use numerical simulation to evaluate the performance of zero-bias DNN under different $Q$ values and different sequential detection algorithms: CUSUM\cite{granjon2013cusum}, EWMA (Exponentially Weighted Moving Average \cite{1179803}) and sliding window\cite{10.1145/502585.502630}. We simulate the possible values of $h_{GLR}$, $FPR$, and $TPR$ that a binary abnormality detector can encounter with $TPR \in [0.6,0.99]$, $FPR=0.4$, $Q \in [1.625, 2.25]$. We configure three sequential detection algorithms as follows:
	\begin{itemize}
		\item \textit{CUSUM: }we set the event detection threshold $h_{CUSUM} \in [2.0,20.0]$.
		\item \textit{EWMA: }we set $\lambda = 0.15$ and $L \in [3.0,4.0]$.
		\item \textit{Sliding window: }we set the length of window $L \in [50, 300]$ with a threshold 0.7.
	\end{itemize}
	We first compare the best and the worst detection delay of the three sequential event detection methods in Figure~\ref{figCompDetectionDelayAll}. We found that considering the best case, the detection delays of EWMA and CUSUM algorithms are close while in the worst case the CUSUM algorithm performs better than EWMA and sliding window.
	The averaged detection delays as well as its range are compared in Figure~\ref{figAvgDetectionDelayAllCases} and \ref{figCompDetectionDelayRangeAll}. Although EWMA algorithm seems to achieve the best performance in the averaged detection delay, the ranges of detection delay in Figure~\ref{figCompDetectionDelayRangeAll} reveal that the EWMA algorithm is not very stable when the $Q$ value is not sufficiently large. As predicted, the sliding window algorithm always has the worst performance.
	\subsection{Incremental learning}
	To evaluate our incremental learning mechanism, we separate our data set into two parts, namely \textit{task-1} and \textit{task-2}, respectively. We first train the zero-bias DNN on \textit{task-1} and use continual learning mechanisms to let our network recognize wireless transmitters in \textit{task-2}.
	
	\subsubsection{Numerical stability}
\begin{figure}[h]
		\centering
		\includegraphics[width=0.8\linewidth]{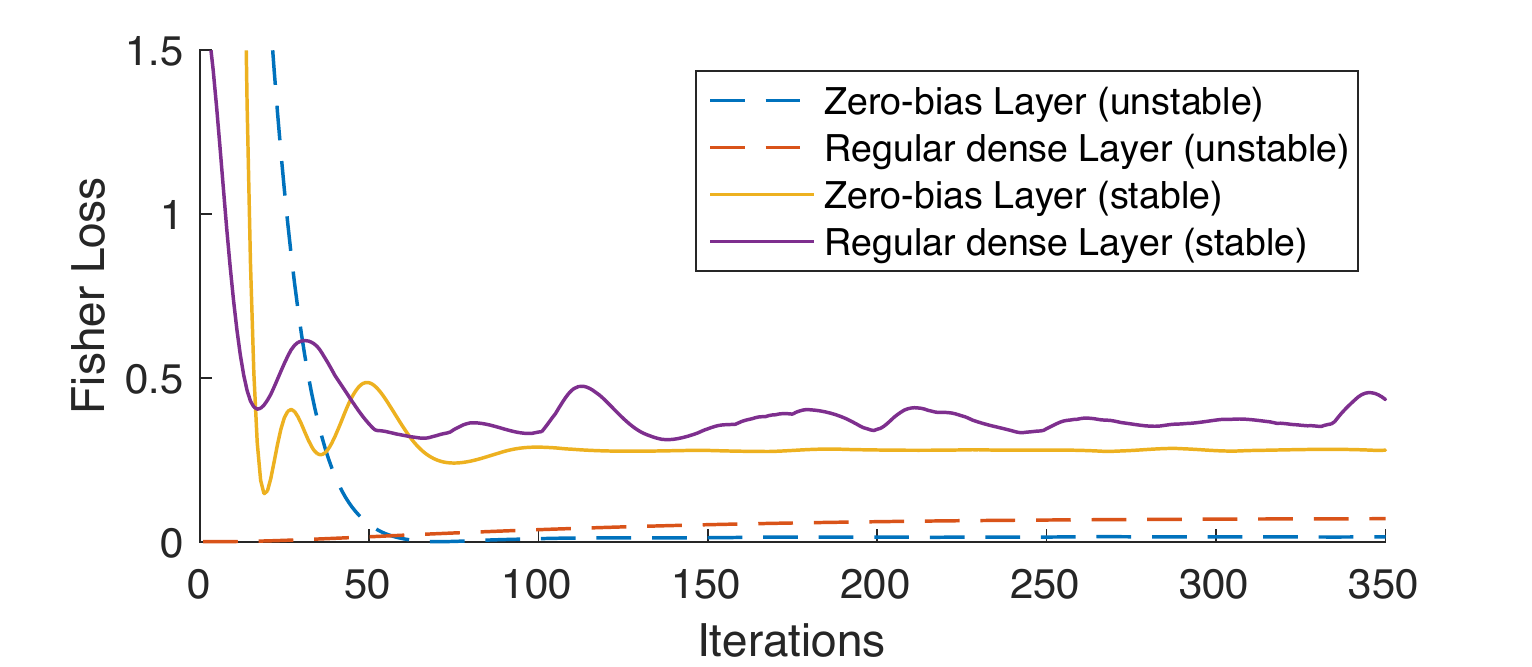}
		\caption{Compare of numerical stability during continual learning}
		\label{figFisherInfoStability}
	\end{figure}

We compare the numerical stability of Fisher Loss during incremental learning. The results in Figure~\ref{figFisherInfoStability} demonstrate that without applying the exponential function as in Equation (\ref{eqFisherExp}), the Fisher Loss is numerically unstable and gradually vanishes to zero (depicted by dashed lines). When Fisher Loss becomes zero, the incremental learning algorithm can no longer penalize the neural network for forgetting the old tasks. In contrast, if the exponential function is applied, the Fisher Loss never vanishes to zero and prevents catastrophic forgetting. As incremental learning procedures, the Fisher Loss gradually converges to a nonzero constant value. The results indicate that the zero-bias layer has a smoother converging characteristic than the regular dense layer. 
	
	\subsubsection{Comparison of incremental learning approaches}
	This subsection will compare other incremental learning approaches with our method. The descriptions of all these approaches are given in Table~\ref{tabContinualLearnApproaches}. We aim to compare the effect of EWC as well as other network knowledge protection methods. Please be noted that during the incremental learning, $L2$ regularization factors for the regular dense layer and zero-bias layer are all set to 0 and 0.025, respectively. 
	\begin{table}[h]
		\centering
		\caption{Compared approaches for continual learning}
		\label{tabContinualLearnApproaches}
		\resizebox{.4\textwidth}{!}{
			\begin{tabular}{@{}llll@{}}
				\toprule
				\multicolumn{1}{c}{Approaches} &
				\multicolumn{1}{c}{\begin{tabular}[c]{@{}c@{}}Lock zero-bias\\ layer\end{tabular}} &
				\multicolumn{1}{c}{\begin{tabular}[c]{@{}c@{}}Elastic Weight \\ Consolidation\end{tabular}} &
				\multicolumn{1}{c}{\begin{tabular}[c]{@{}c@{}}Locked prior \\ layers\end{tabular}} \\ \midrule
				Global EWC &
				No &
				Globally &
				No \\ \midrule
				\begin{tabular}[c]{@{}l@{}}Only train new \\ Fingerprints\end{tabular} &
				\begin{tabular}[c]{@{}l@{}}Lock old \\ fingerprints\end{tabular} &
				No &
				Yes \\ \midrule
				\begin{tabular}[c]{@{}l@{}}Only protect old\\ Fingerprints\end{tabular} &
				\begin{tabular}[c]{@{}l@{}}Lock old \\ fingerprints\end{tabular} &
				Yes &
				No \\ \midrule
				\begin{tabular}[c]{@{}l@{}}Only use EWC in\\ the last layer\end{tabular} &
				No &
				\begin{tabular}[c]{@{}l@{}}Only in the \\last layer.\end{tabular} &
				Yes \\ \bottomrule
			\end{tabular}
		}
	\end{table}
	
	\begin{figure}[t]
		\centering  
		\subfloat[Zero-bias layer]
		{%
			\includegraphics[width=0.8\linewidth]{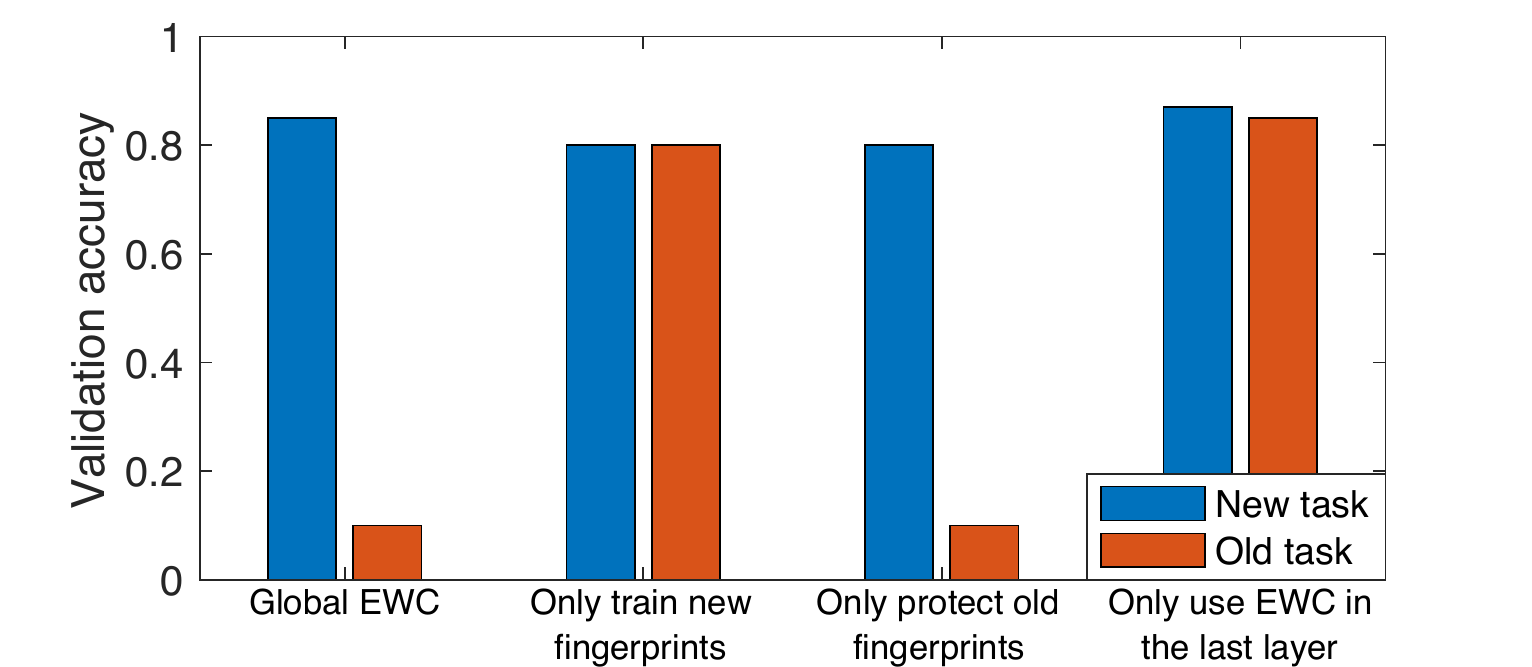}
			\label{figFingerprintCLScenarios}
		}\\
		\subfloat[Regular dense layer]
		{
			\includegraphics[width=0.8\linewidth]{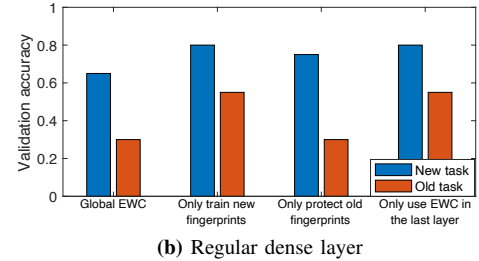}
			\label{figDenseCLScenarios}
		}%
		\caption{Performance comparison of zero-bias layer and regular dense layer DNNs for incremental learning}
		\label{figCLScenarios}
		
	\end{figure}
	
	The results are given in Figure~\ref{figCLScenarios}, with the following highlights:
	\begin{enumerate}
		\item In Global EWC, catastrophic forgetting is not prevented. Besides, the zero-bias layer retains far less knowledge from previous tasks.
		\item Only training new fingerprints and locking all old weights in the network can help retaining knowledge from previous tasks. This phenomenon indicates that the prior layers have already extracted useful features for the final classification. Moreover, the zero-using bias layer's performance indicates that it can enable prior neural network layers to discover better features without relying on biases and weights. Please be noted that this scenario also prevents the fine-tuning of existing fingerprints even if they are in sub-optimal directions.
		\item Only protecting old fingerprints does not seem to be helpful. The new task will destroy all useful feature extractors in prior layers.
		\item Applying EWC only in the last layer provides the most promising results. Notably, the neural networks with the zero-bias layer still outperform regular neural networks. This fact explains that EWC tries to protect old fingerprints from changing erroneously (forgetting) and enables fine-tuning.
	\end{enumerate}
	
	\section{Conclusion}
	\label{sectCC}
	This paper significantly extends the analysis of our previously proposed zero-bias DNN and combines it with the Quickest Detection algorithms to detect abnormalities and time-dependent abnormal events in IoT with the lowest assured latency. We first analyze the zero-bias DNN and show that zero-bias DNN is superior to regular DNN for RF signal surveillance. We then propose a novel solution to convert zero-bias DNN classifiers into performance-assured binary abnormality detectors. We model the converted abnormality detectors using Bernoulli distribution, which perfectly adapts to the CUSUM-based Quickest Detection scheme. The theoretically assured lowest abnormal event detection delay is provided with predictable false alarms in this Quickest Detection scheme. Finally, to facilitate DNN for RF signal surveillance to evolve incrementally, we propose a more stable EWC algorithm and shown that zero-bias DNN is more reliable than regular DNN under incremental learning. The framework is evaluated using both massive signal records from real-world aviation communication systems and simulated data. In the future, we will explore the incremental learning capability of zero-bias DNN.
	\section*{Acknowledgment}
	
	This research was partially supported by the Center for Advanced Transportation Mobility (CATM), USDOT under Grant No. 69A3551747125 and the National Science Foundation under Grant No. 1956193.
	
	\bibliographystyle{IEEEtran}
	\bibliography{ReviewRef.bib}

\begin{thebibliography}{10}
\providecommand{\url}[1]{#1}
\csname url@samestyle\endcsname
\providecommand{\newblock}{\relax}
\providecommand{\bibinfo}[2]{#2}
\providecommand{\BIBentrySTDinterwordspacing}{\spaceskip=0pt\relax}
\providecommand{\BIBentryALTinterwordstretchfactor}{4}
\providecommand{\BIBentryALTinterwordspacing}{\spaceskip=\fontdimen2\font plus
\BIBentryALTinterwordstretchfactor\fontdimen3\font minus
  \fontdimen4\font\relax}
\providecommand{\BIBforeignlanguage}[2]{{%
\expandafter\ifx\csname l@#1\endcsname\relax
\typeout{** WARNING: IEEEtran.bst: No hyphenation pattern has been}%
\typeout{** loaded for the language `#1'. Using the pattern for}%
\typeout{** the default language instead.}%
\else
\language=\csname l@#1\endcsname
\fi
#2}}
\providecommand{\BIBdecl}{\relax}
\BIBdecl

\bibitem{7406686}
Y.~{Sun}, H.~{Song}, A.~J. {Jara}, and R.~{Bie}, ``Internet of things and big
  data analytics for smart and connected communities,'' \emph{IEEE Access},
  vol.~4, pp. 766--773, 2016.

\bibitem{song2016cyber}
H.~Song, D.~B. Rawat, S.~Jeschke, and C.~Brecher, \emph{Cyber-physical systems:
  foundations, principles and applications}.\hskip 1em plus 0.5em minus
  0.4em\relax Elsevier, 2016.

\bibitem{9345787}
L.~Wang, W.~Song, Y.~Lan, H.~Wanga, X.~Yue, X.~Yin, E.~Luo, B.~Zhang, Y.~Lu,
  and Y.~Tang, ``A smart droplet detection approach with vision sensing
  technique for agricultural aviation application,'' \emph{IEEE Sensors
  Journal}, pp. 1--1, 2021.

\bibitem{8897627}
I.~{Butun}, P.~{Österberg}, and H.~{Song}, ``Security of the internet of
  things: Vulnerabilities, attacks, and countermeasures,'' \emph{IEEE
  Communications Surveys Tutorials}, vol.~22, no.~1, pp. 616--644, 2020.

\bibitem{9500733}
J.~Wang, Y.~Liu, S.~Niu, and H.~Song, ``Reinforcement learning optimized
  throughput for 5g enhanced swarm uas networking,'' in \emph{ICC 2021 - IEEE
  International Conference on Communications}, 2021, pp. 1--6.

\bibitem{9425491}
Y.~Liu, J.~Wang, J.~Li, S.~Niu, and H.~Song, ``Class-incremental learning for
  wireless device identification in iot,'' \emph{IEEE Internet of Things
  Journal}, pp. 1--1, 2021.

\bibitem{liu2021machine}
------, ``Machine learning for the detection and identification of internet of
  things (iot) devices: A survey,'' 2021.

\bibitem{wang2021counter}
J.~Wang, Y.~Liu, and H.~Song, ``{Counter-Unmanned Aircraft System (s)(C-UAS):
  State of the Art, Challenges, and Future Trends},'' \emph{IEEE Aerospace and
  Electronic Systems Magazine}, vol.~36, no.~3, pp. 4--29, 2021.

\bibitem{jiang2019uncertainty}
Y.~Jiang, M.~Wang, X.~Jiao, H.~Song, H.~Kong, R.~Wang, Y.~Liu, J.~Wang, and
  J.~Sun, ``Uncertainty theory based reliability-centric cyber-physical system
  design,'' in \emph{2019 International Conference on Internet of Things
  (iThings) and IEEE GreenCom/CPSCom/SmartData}.\hskip 1em plus 0.5em minus
  0.4em\relax IEEE, 2019, pp. 208--215.

\bibitem{perera2017efficient}
P.~Perera and V.~M. Patel, ``Efficient and low latency detection of intruders
  in mobile active authentication,'' \emph{IEEE Transactions on Information
  Forensics and Security}, vol.~13, no.~6, pp. 1392--1405, 2017.

\bibitem{liu2020zero}
Y.~Liu, J.~Wang, J.~Li, H.~Song, T.~Yang, S.~Niu, and Z.~Ming, ``Zero-bias deep
  learning for accurate identification of internet-of-things (iot) devices,''
  \emph{IEEE Internet of Things Journal}, vol.~8, no.~4, pp. 2627--2634, 2021.

\bibitem{scheirer2012toward}
W.~J. Scheirer, A.~de~Rezende~Rocha, A.~Sapkota, and T.~E. Boult, ``Toward open
  set recognition,'' \emph{IEEE transactions on pattern analysis and machine
  intelligence}, vol.~35, no.~7, pp. 1757--1772, 2012.

\bibitem{bendale2016towards}
A.~Bendale and T.~E. Boult, ``Towards open set deep networks,'' in
  \emph{Proceedings of the IEEE conference on computer vision and pattern
  recognition}, 2016, pp. 1563--1572.

\bibitem{wong2018clustering}
L.~J. Wong, W.~C. Headley, S.~Andrews, R.~M. Gerdes, and A.~J. Michaels,
  ``{Clustering learned CNN features from raw I/Q data for emitter
  identification},'' in \emph{MILCOM 2018-2018 IEEE Military Communications
  Conference (MILCOM)}.\hskip 1em plus 0.5em minus 0.4em\relax IEEE, 2018, pp.
  26--33.

\bibitem{shi2019deep}
Y.~Shi, K.~Davaslioglu, Y.~E. Sagduyu, W.~C. Headley, M.~Fowler, and G.~Green,
  ``{Deep Learning for RF Signal Classification in Unknown and Dynamic Spectrum
  Environments},'' in \emph{2019 IEEE International Symposium on Dynamic
  Spectrum Access Networks (DySPAN)}.\hskip 1em plus 0.5em minus 0.4em\relax
  IEEE, 2019, pp. 1--10.

\bibitem{roy2019rfal}
D.~Roy, T.~Mukherjee, M.~Chatterjee, E.~Blasch, and E.~Pasiliao, ``Rfal:
  Adversarial learning for rf transmitter identification and classification,''
  \emph{IEEE Transactions on Cognitive Communications and Networking}, 2019.

\bibitem{lai2008quickest}
L.~Lai, Y.~Fan, and H.~V. Poor, ``{Quickest Detection in Cognitive Radio: A
  Sequential Change Detection Framework},'' in \emph{IEEE GLOBECOM 2008 - 2008
  IEEE Global Telecommunications Conference}, 2008, pp. 1--5.

\bibitem{poor2008quickest}
H.~Poor and O.~Hadjiliadis, \emph{\BIBforeignlanguage{English (US)}{Quickest
  detection}}.\hskip 1em plus 0.5em minus 0.4em\relax United Kingdom: Cambridge
  University Press, Jan. 2008, vol. 9780521621045.

\bibitem{johnson2017detecting}
\BIBentryALTinterwordspacing
P.~Johnson, J.~Moriarty, and G.~Peskir, ``Detecting changes in real-time data:
  a users' guide to optimal detection,'' \emph{Philosophical Transactions of
  the Royal Society A: Mathematical, Physical and Engineering Sciences}, vol.
  375, no. 2100, p. 20160298, 2017. [Online]. Available:
  \url{https://royalsocietypublishing.org/doi/abs/10.1098/rsta.2016.0298}
\BIBentrySTDinterwordspacing

\bibitem{basseville1993detection}
M.~Basseville, I.~V. Nikiforov \emph{et~al.}, \emph{Detection of abrupt
  changes: theory and application}.\hskip 1em plus 0.5em minus 0.4em\relax
  prentice Hall Englewood Cliffs, 1993, vol. 104.

\bibitem{granjon2013cusum}
P.~Granjon, ``The {CuSum} algorithm-a small review,'' 2013.

\bibitem{wang2018deep}
M.~Wang and W.~Deng, ``Deep visual domain adaptation: A survey,''
  \emph{Neurocomputing}, vol. 312, pp. 135--153, 2018.

\bibitem{matlabMinst}
``Create simple deep learning network for classification - \text{MATLAB} \&
  \text{S}imulink example,''
  \url{https://www.mathworks.com/help/deeplearning/ug/create-simple-deep-learning-network-for-classification.html},
  (Accessed on 07/12/2021).

\bibitem{maaten2008visualizing}
L.~v.~d. Maaten and G.~Hinton, ``{Visualizing data using t-SNE},''
  \emph{Journal of machine learning research}, vol.~9, no. Nov, pp. 2579--2605,
  2008.

\bibitem{gt9v-kz32-20}
\BIBentryALTinterwordspacing
Y.~Liu, J.~Wang, H.~Song, S.~Niu, and Y.~Thomas, ``A 24-hour signal recording
  dataset with labels for cybersecurity and \text{IoT},'' 2020. [Online].
  Available: \url{http://dx.doi.org/10.21227/gt9v-kz32}
\BIBentrySTDinterwordspacing

\bibitem{liu2020deep}
Y.~Liu, J.~Wang, S.~Niu, and H.~Song, ``Deep learning enabled reliable identity
  verification and spoofing detection,'' in \emph{Wireless Algorithms, Systems,
  and Applications}, D.~Yu, F.~Dressler, and J.~Yu, Eds.\hskip 1em plus 0.5em
  minus 0.4em\relax Cham: Springer International Publishing, 2020, pp.
  333--345.

\bibitem{weisstein2002bernoulli}
E.~W. Weisstein, ``Bernoulli distribution,'' \emph{https://mathworld. wolfram.
  com/}, 2002.

\bibitem{xie2021sequential}
L.~Xie, S.~Zou, Y.~Xie, and V.~V. Veeravalli, ``{Sequential (Quickest) Change
  Detection: Classical Results and New Directions},'' \emph{IEEE Journal on
  Selected Areas in Information Theory}, vol.~2, no.~2, pp. 494--514, 2021.

\bibitem{kirkpatrick2017overcoming}
J.~Kirkpatrick, R.~Pascanu, N.~Rabinowitz, J.~Veness, G.~Desjardins, A.~A.
  Rusu, K.~Milan, J.~Quan, T.~Ramalho, A.~Grabska-Barwinska \emph{et~al.},
  ``Overcoming catastrophic forgetting in neural networks,'' \emph{Proceedings
  of the national academy of sciences}, vol. 114, no.~13, pp. 3521--3526, 2017.

\bibitem{riddle1090}
J.~Sun, ``{An open-access book about decoding Mode-S and ADS-B data},''
  \url{https://mode-s.org/}, May 2017.

\bibitem{1179803}
N.~Ye, S.~Vilbert, and Q.~Chen, ``Computer intrusion detection through ewma for
  autocorrelated and uncorrelated data,'' \emph{IEEE Transactions on
  Reliability}, vol.~52, no.~1, pp. 75--82, 2003.

\bibitem{10.1145/502585.502630}
\BIBentryALTinterwordspacing
C.-H. Lee, C.-R. Lin, and M.-S. Chen, ``Sliding-window filtering: An efficient
  algorithm for incremental mining,'' ser. CIKM '01.\hskip 1em plus 0.5em minus
  0.4em\relax New York, NY, USA: Association for Computing Machinery, 2001, p.
  263–270. [Online]. Available: \url{https://doi.org/10.1145/502585.502630}
\BIBentrySTDinterwordspacing

\end{thebibliography}

\end{document}